# Interactive Design by Integrating a Large Pre-Trained Language Model and Building Information Modeling

Suhyung Jang[1] Ghang Lee, Ph.D.[2]

[1]Building Informatics Group, Dept. of Architecture and Architectural Engineering, Yonsei Univ. ORCID: https://orcid.org/ 0000-0001-8289-0657. Email: rgb000@yonsei.ac.kr.
[2]Professor, Dept. of Architecture and Architectural Engineering, Yonsei Univ. (corresponding author). ORCID: https://orcid.org/ 0000-0002-3522-2733. Email: glee@yonsei.ac.kr

## ABSTRACT

This study explores the potential of generative artificial intelligence (AI) models, specifically OpenAI's generative pre-trained transformer (GPT) series, when integrated with building information modeling (BIM) tools as an interactive design assistant for architectural design. The research involves the development and implementation of three key components: 1) BIM2XML, a component that translates BIM data into extensible markup language (XML) format; 2) Generative AI-enabled Interactive Architectural design (GAIA), a component that refines the input design in XML by identifying designer intent, relevant objects, and their attributes, using pre-trained language models; and 3) XML2BIM, a component that converts AI-generated XML data back into a BIM tool. This study validated the proposed approach through a case study involving design detailing, using the GPT series and Revit. Our findings demonstrate the effectiveness of state-of-the-art language models in facilitating dynamic collaboration between architects and AI systems, highlighting the potential for further advancements.

## INTRODUCTION

The architectural design process is complex, demanding the incorporation of diverse ideas, constraints, and stakeholders to produce innovative and functional solutions. Traditional design methods often involve labor-intensive processes and limited productivity (Eldin 1991). Building information modeling (BIM) has increasingly been adopted to streamline the building project life cycle, including design, construction, and facility management phases (Sacks et al. 2018). Additionally, researchers have begun to harness BIM models and associated data using artificial intelligence (AI) technology (Pan and Zhang 2022). However, the substantial untapped potential remains for enhancing BIM and AI integration to lower the technical barriers limiting BIM adoption (Won et al. 2013).

Large pre-trained language models, such as OpenAI's generative pre-trained transformer (GPT) series, demonstrate remarkable capabilities in understanding and generating human-like text. These AI models can analyze vast amounts of data, recognize patterns, and generate meaningful insights, making them a promising tool for transforming the architectural design process. By incorporating these state-of-the-art language models into the BIM authoring process, architects can benefit from dynamic collaboration with AI systems, enhancing both the efficiency and quality of their design outcomes.

This study aims to explore the potential of integrating large pre-trained language models with BIM to revolutionize the interactive design process in architecture. The research focuses on the development of three key components: BIM2XML, Generative AI-enabled Interactive Architectural design (GAIA), and XML2BIM, and validation through a case study.

The paper is structured as follows: a literature review on BIM and AI language models in architectural design, a detailed research method, a presentation of the case study, results and discussion, and a conclusion including implications of the findings for future research and advancements in the field.

**LITERATURE REVIEW**

BIM has emerged as a powerful tool for architects, enabling the digital representation of physical and functional characteristics of buildings throughout their lifecycle (Sacks et al. 2018), improving collaboration, streamlining workflows, and reducing errors in the design process. However, despite its growing popularity and adoption, BIM still faces challenges such as interoperability, standardization, and the need for continuous improvement (Won et al. 2013).

Recent advancements in AI, particularly in natural language processing (NLP), have resulted in the development of large pre-trained language models (Brown et al. 2020; Radford et al. 2018). These models have demonstrated remarkable capabilities in generating human-like text, understanding context, and providing valuable insights across various domains. A recent technical report indicates that the performance of extremely large-scale language models can outperform models trained for specific uses (OpenAI 2023).

The integration of AI techniques into the architectural design process has garnered significant attention in recent years (Pan and Zhang 2022). Research has explored various AI applications, such as generative design (Qian et al. 2022), design optimization (Salehi and Burgueño 2018), and automated code compliance (Sacks et al. 2019). However, the integration of AI language models with BIM to facilitate interactive design processes remains an underexplored area of research.

There also have been efforts to explore the potential of combining AI language models with BIM. Wu et al. (Wu et al. 2019) developed a natural language-based retrieval engine for BIM object databases, while Elghaish et al. (Elghaish et al. 2022) proposed a solution using an AI voice assistant platform to provide remote interaction and retrieve information from a BIM model. These studies provide a foundation for the potential benefits of integrating AI language models with BIM. However, there remains a gap in previous studies regarding the integration of AI models with BIM to streamline the interactive design process by connecting the design intent and functions of BIM authoring tools. In this regard, the current research was conducted to address the current gap by proposing the implementation of a framework that combines large pre-trained language models with BIM for human-AI interaction.

**RESEARCH METHOD**

The framework proposed in this study consists of three main components: 1) BIM2XML; 2) GAIA; and 3) XML2BIM. Each component was developed using the C# application programming interface (API) of Revit and the GPT series without fine-tuning.

BIM2XML is a component that transforms building elements represented in Revit into extensible markup language (XML) format. XML is commonly used and easily processed by language models trained on vast amounts of data. Instead of using the Industry Foundation Classes (IFC) format, which is based on the Standard for the Exchange of Product (STEP) and has token limitations, XML enables a more efficient and agile representation of building elements with only the necessary attributes for a given task.

GAIA is a component for elaborating input XML to apply design tasks given by the users using pre-trained language models. GAIA appends predefined instructions to the user's prompt to guide the model in generating output XML that reflects the designer's intent, considering relevant objects and attribute changes. The authors manually copied and pasted the BIM2XML output, then appended the design task prompt along with the predefined instruction. For this study, GAIA was manually implemented as a proof of concept using OpenAI playground and ChatGPT, though future research could automate this process using the GPT API.

XML2BIM is a component for converting the output XML back into a BIM model, facilitating the integration of AI-supported design suggestions into the BIM authoring environment. This component identifies the changes made using GAIA and applies corresponding changes in the BIM models using the Revit C# API.

To assess the effectiveness of the proposed framework, a case study was conducted that involved wall detailing tasks. The performance of the framework was evaluated using commonly used metrics for assessing classification model performance, such as accuracy, precision, recall, and F1-score. The models evaluated in this study were GPT-3, GPT-3.5, and GPT-4 demo. These metrics were calculated based on the most frequent answers from five iterative predictions provided by the three language models. The purpose of this evaluation was to compare the performance of the proposed framework across different models. Accuracy quantifies the proportion of correct predictions made by the model, while precision measures the ratio of true positives to all positive predictions. Recall, on the other hand, calculates the ratio of true positives to all actual positive cases. The F1-score is a harmonic mean of precision and recall and provides a balanced measure of the model's overall performance. The results of the evaluation were visualized as a comparison matrix to identify in which specific task each model performed well or was confused.
Furthermore, a multi-rater Cohen's Kappa algorithm was developed to assess the consistency of the models in providing answers across five iterative, identical tasks. The dataset, stored in a CSV file, included predictions made by the framework during the five iterations. The algorithm extracted the predicted labels from the CSV file and computed the multi-rater Cohen's Kappa coefficient by constructing a contingency table that represented the number of raters who assigned each label to each instance. This metric offered a comprehensive assessment of model performance in terms of inter-rater agreement, accounting for consistency of the model. Therefore, it was suitable for evaluating natural language models that generate heuristic answers.

**CASE STUDY**

A case study was conducted to evaluate the performance of the proposed approach, focusing on the wall detailing process during the design development phase. Wall detailing is a critical aspect of architectural design, involving material specification, finishes, and construction methods that significantly impact building performance and aesthetics.

Figure 1 depicts the flow of the case study. For this study, a schematic BIM model of Villa Savoye and a detailed version of the BIM model were prepared as the golden standard. The schematic model had rooms tagged by the names of the programs, and the wall types were included as libraries but not applied as wall details. Even though room names were used as tags, architects can easily identify "Master bedroom," "Bedroom," "Ramp," "Hallway," and "Private sitting room" as indoor spaces; "Terrace" and "Kitchen terrace" as outdoor spaces; and "Kitchen," "Bathroom," and "Toilet" as wet rooms. The authors refined the detailed BIM model by considering different spatial characteristics (e.g., indoors, outdoors, and wet rooms) present on each side of the walls. In the model, there are five cases: outdoor space meets indoor space ("EIFS on Mtl. Stud with gypsum finish 300mm"), indoor space meets indoor space ("Gypsum finishes 150mm"), wet room meets outdoor space ("EIFS on Mtl. Stud with tile finish 300mm"), wet room meets indoor space ("Gypsum and tile finish 150mm"), and wet room meets wet room ("Tile finishes 150mm"). The validation scope aimed to evaluate whether the language model could understand and apply the implicit logic of understanding architectural context in wall detailing, given the XML-formatted BIM model and names of the details used for the detailing task.
The validation aimed to assess whether the language model could comprehend and apply the implicit logic of understanding architectural context in wall detailing hidden in the guideline. The process involved translating the attributes, topology, and context of 48 wall elements from the case models into XML, prompting the language models using GAIA, and generating new XML files containing the required wall details. The XML2BIM algorithm then applied these changes to the BIM model. The prompt was tested in five iterations on three language models: GPT-3, GPT-3.5, and GPT-4 demo, with the results compared to the golden standard. The case study outcomes and an analysis of the performance are presented in the following section.

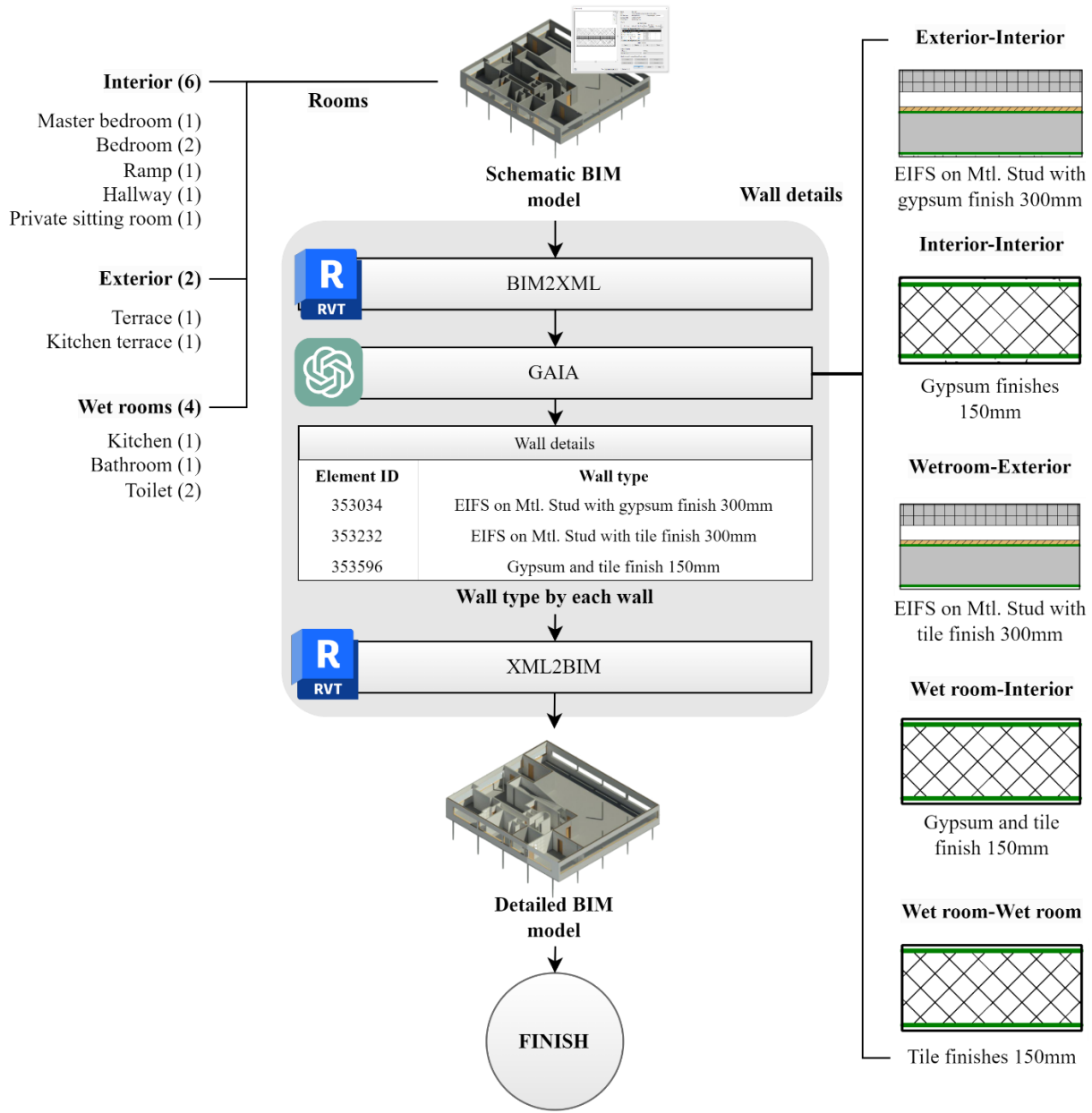

**Figure 1. Flowchart of the case study on the proposed method.**

**RESULTS AND DISCUSSION**

Table 1 displays the performance metrics of GPT-3, GPT-3.5, and GPT-4 demo on the proposed framework. GPT-4 demo achieved the best performance, with an accuracy score of 0.83, precision score of 0.64, recall score of 0.60, and an F1-score of 0.62. In comparison, GPT-3.5 had an accuracy score of 0.70, precision score of 0.55, recall score of 0.56, and an F1-score of 0.55, while GPT-3 displayed the lowest performance with an accuracy score of 0.51, precision score of 0.47, recall score of 0.44, and an F1-score of 0.40.

**Table 1. Comparison of Performance Metrics for GPT-3, GPT-3.5, and GPT-4 demo.**

| Performance metrics | GPT-3 | GPT-3.5 | GPT-4 demo |
|---|---|---|---|
| Accuracy | 0.51 | 0.70 | 0.83 |
| Precision | 0.47 | 0.55 | 0.64 |
| Recall | 0.44 | 0.56 | 0.60 |
| F1-score | 0.40 | 0.55 | 0.62 |

To further analyze each model's performance on each wall detail, a confusion matrix was generated and is presented in Figure 2. The x-axis and y-axis of the confusion matrix, labeled from 0 to 5, correspond to the following order: "Generic – 150mm", "Tile finishes 150mm", "EIFS on Mtl. Stud with tile finish 300mm", "Gypsum and tile finish 150mm", "EIFS on Mtl. Stud with gypsum finish 300mm", and "Gypsum finishes 150mm". The results revealed that GPT-3 had difficulty applying changes according to the guidelines, leaving one "Generic - 150mm" wall unmodified. GPT-3.5 exhibited slight confusion when handling wet rooms. In contrast, GPT-4 demo performed well for most wall types but failed to apply the "Tile finishes 150mm" modification, which was present only once in the model. Overall, GPT-4 demo demonstrated better performance compared to GPT-3 and GPT-3.5, as evidenced by higher values across all metrics.

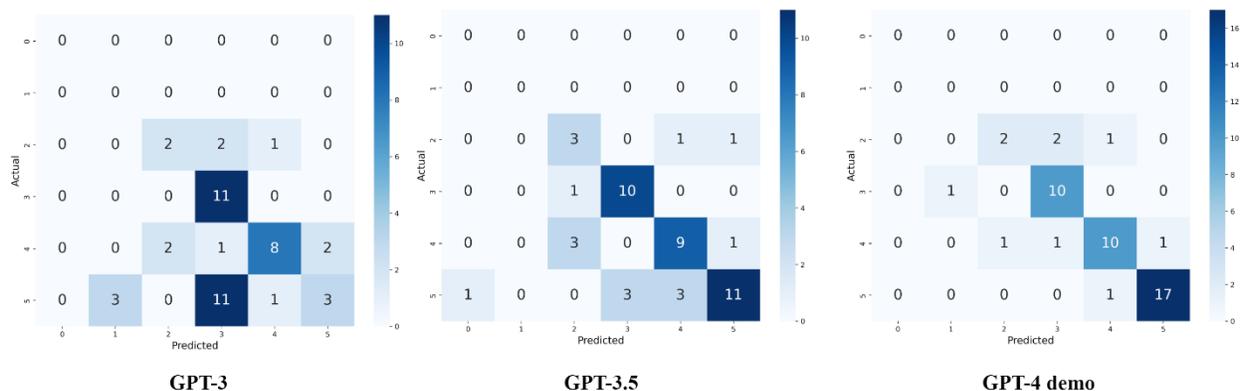

**Figure 2. Confusion matrix of GPT-3, GPT-3.5, and GPT-4 demo.**

Figure 3 illustrates a comparison between the original model (a), the model refined using GPT-3 (b), and the model refined with GPT-4 demo (c). Upon closer examination of the detailed models for GPT-3 (d) and GPT-4 demo (e), it becomes evident that the intersections between wet rooms and interior spaces are more effectively managed in the GPT-4 demo, compared to GPT-3.

The authors also assessed the framework using the multi-rater Cohen's kappa coefficient. A higher kappa coefficient signifies a higher level of agreement between GAIA's answers on repeated interactives. Cohen suggested the Kappa result be interpreted as follows (McHugh 2012): 0.00–0.20 indicating no agreement, 0.21–0.39 as minimal, 0.40-0.59 as weak, 0.60-0.79 as moderate, 0.80-0.90 as strong, and above 0.90 as almost perfect. Table 2 shows the multi-rater Cohen's kappa coefficient values for GPT-3, GPT-3.5, and GPT-4 across five different wall types, indicating the agreement level between the five interactives of each model's predictions on each wall type.

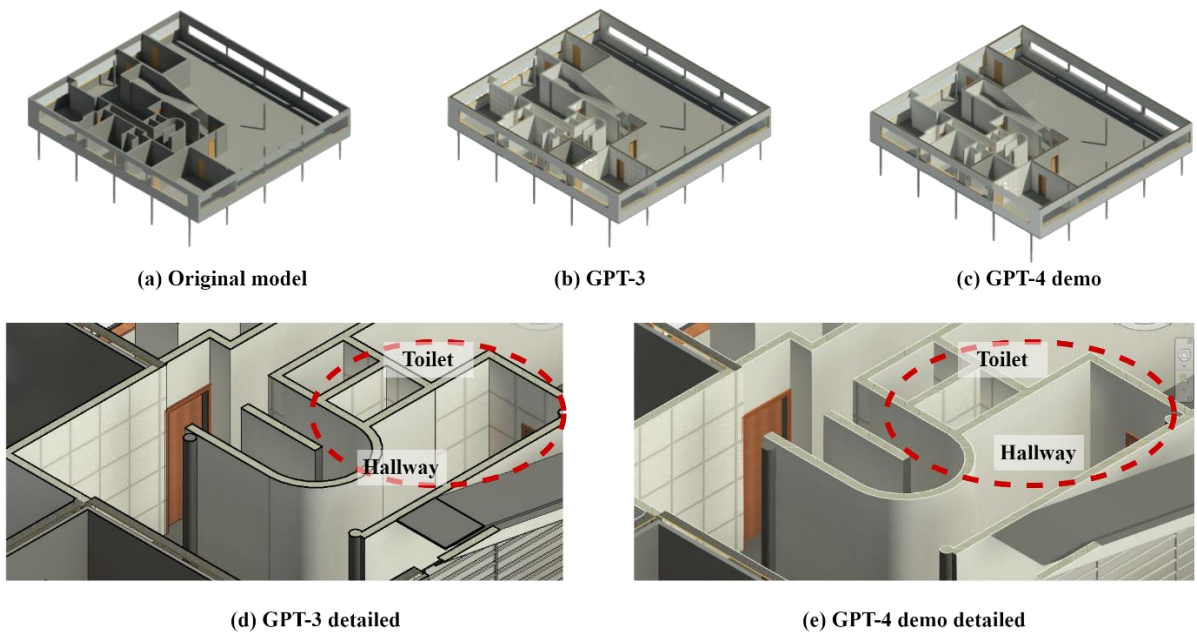

(a) Original model　　　　　(b) GPT-3　　　　　(c) GPT-4 demo

(d) GPT-3 detailed　　　　　(e) GPT-4 demo detailed

**Figure 3. Comparison of detailed results of GPT-3 and GPT-4 demo.**

GPT-3 achieved a moderate kappa coefficient value for "EIFS on Mtl. Stud with gypsum finish 300mm", but generally exhibited weaker consistent answers compared to the other two models, arranging from no agreement to weak agreement. GPT-3.5 achieved strong kappa coefficient values for "Gypsum and tile finish 150mm", "Gypsum finishes 150mm", and "EIFS on Mtl. Stud with tile finish 300mm", indicating the strongest agreement between its predictions existed. In addition, GPT-3.5 achieved a moderate kappa coefficient value of 0.76 for "EIFS on Mtl. Stud with gypsum finish 300mm", which was the highest value among all models. However,

GPT-3.5 achieve a weak kappa coefficient value for "EIFS on Mtl. Stud with tile finish 300mm". GPT-4 demo achieved moderate kappa coefficient values ranging from 0.61 to 0.71 for overall wall types, except for the "Tile finishes 150mm" which only exists once in the model.

**Table 2. Multi-rater Cohen's kappa coefficient of GPT-3, GPT-3.5, and GPT-4 demo by wall types.**

| Wall types | GPT-3 | GPT-3.5 | GPT-4 demo |
|---|---|---|---|
| Gypsum and tile finish 150mm | 0.06 | 0.86 | 0.61 |
| Gypsum finishes 150mm | 0.51 | 0.82 | 0.68 |
| Tile finishes 150mm | 0.00 | 0.00 | 0.00 |
| EIFS on Mtl. Stud with tile finish 300mm | 0.45 | 0.49 | 0.65 |
| EIFS on Mtl. Stud with gypsum finish 300mm | 0.70 | 0.76 | 0.71 |

Considering the result from performance evaluation and detailing results, it can be concluded that GPT-4 demo demonstrated the best performance among the three language models. Moreover, the multi-rater Cohen's kappa coefficient results showed that GPT-3 achieved the weakest agreement among the three language models, while GPT-3.5 and GPT-4 demo showed moderate or strong agreement for most of the wall types. In conclusion, the proposed framework, when tested with GPT-4 demo, successfully demonstrated the ability to comprehend and apply the implicit logic of understanding the architectural context in wall detailing. The results indicate that the integration of advanced language models with BIM can potentially enhance the efficiency and accuracy of the architectural design process. However, it is important to recognize the limitations of the current study, including the small sample size and the focus on a single case study. Further research should explore the generalizability of these findings across a wider range of architectural projects and applications, as well as the potential for continuous improvement as language models evolve.

**CONCLUSION AND FUTURE WORK**

This study highlights the potential of integrating large pre-trained language models with BIM authoring tools to metamorphose the architectural design process. By developing and implementing BIM2XML, GAIA, and XML2BIM using the GPT series and Revit C# API, the authors have established a framework that facilitates collaboration between architects and AI language models during the design process. The case study on wall detailing yielded promising results, with AI models providing consistent and context-aware suggestions that contribute to efficient collaboration with architects, improving productivity, and reducing the technical barriers to BIM adoption.

Although there are limitations and areas for improvement, the findings of this study underscore the potential for further advancements in integrating AI language models with BIM authoring tools. Future research can focus on refining and enhancing GAIA, broadening the scope of design tasks to better understand the potential of AI language models across various aspects of architectural design, and developing user interfaces and visualization tools to enable more intuitive communication between architects and AI models. By continuing to explore and refine this approach while addressing these areas, the technical barriers that hinder BIM adoption can be lowered, leading to improved design productivity. Consequently, architects can focus more on showcasing their creativity.

## ACKNOWLEDGMENTS

This work is supported in 2023 by the Korea Agency for Infrastructure Technology Advancement(KAIA) grant funded by the Ministry of Land, Infrastructure and Transport (Grant RS-2021-KA163269).